\def\year{2020}\relax
\documentclass[letterpaper]{article} 
\usepackage{aaai20}  
\usepackage{times}  
\usepackage{helvet} 
\usepackage{courier}  
\usepackage[hyphens]{url}  
\usepackage{graphicx} 
\usepackage{graphicx}  
\usepackage{algorithm}
\usepackage{algorithmic}
\renewcommand{\algorithmicrequire}{\textbf{Input:}} 

\usepackage{subcaption}
\usepackage{caption}
\usepackage[normalem]{ulem}
\usepackage{graphics}

\usepackage{amsmath}
\DeclareMathOperator*{\argmax}{arg\,max}

\usepackage{bm}
\usepackage{amssymb}

\usepackage{natbib}

\newcommand{\R}{\mathbb{R}}
\newcommand{\Rd}{\mathbb R^{d}}
\newcommand{\Rm}{\mathbb R^{m}}

\title{Adversarial Attack with Pattern Replacement}
\author{Ziang Dong, Liang Mao, Shiliang Sun \\ School of Computer Science and Technology, East China Normal University,\\
 3663 North Zhongshan Road, Shanghai 200062, PR China\\ ziangdong@qq.com, lmao14@outlook.com, slsun@cs.ecnu.edu.cn}
\begin{document}
	
	\maketitle
	
	\begin{abstract}
	We propose a generative model for adversarial attack. The model generates subtle but predictive patterns from the input. To perform an attack, it replaces the patterns of the input with those generated based on examples from some other class. We demonstrate our model by attacking CNN on MNIST.
	\end{abstract}
	
	\section{Introduction}
	Recent researches show that machine learning models are vulnerable to adversarial attacks \cite{szegedy2014intriguing, goodfellow2015explaining}. Slightly modifications on input data can fool a state-of-the-art classifier. The adversarial brittleness restricts applications of machine learning models in high security fields, and thus both adversarial attacks and defenses have attracted significant attention these years.
	
	Based on the amount of knowledge the adversary has about the target model, adversarial attacks can be categorized into white box attacks and black box attacks \citep{kurakin2018adversarial}. In white box scenarios, the adversary has all the knowledge about the target model, including the model architecture and all the parameter values, while in the black box scenarios, the adversary can only ``query'' the target model with input data to obtain the output, or even is not allowed to do this.
	
	Most white box attacks generate adversarial examples by directly performing optimization in input space under some norm constraints encouraging visual realism. Instead, we consider to generate the perturbations through a generative model. It is observed that unrecognizable images which do not resemble images from the training set but which typically look like noise while still being classified by the model with high confidence \cite{schott2019towards}. Inspired by this observation, we propose to use a generative model to generate a subtle but predictive patterns from the raw input, and construct adversarial examples by replacing the input's predictive patterns with those from another class.
	
	\section{Methodology}
	
	\paragraph{Problem Definition} Given a target classifier $y=f(x)$, which make decision based on the conditional density of $y$ given $x$, i.e.,
	\begin{equation}
	f(x)=\argmax_y \Tilde{p}(y|x).
	\end{equation}
	One example of such a conditional density is a deep neural network with softmax as the output function of the last layer.
	We want to fool the classifier $f(x)$ by slightly perturbing the input. Instead of learning the perturbations by performing optimization in the input space, we propose to extract the subtle but predictive patterns from the input $x$ and replace them with the patterns extracted from some example of another class. In particular, We model the predictive patterns with a generative model.
	
	\vspace{-3mm}
	\paragraph{Model} Given a labeled example $(x,y)\in\Rd\times\R$, we associate it with a random variable $\eta\in \Rd$. We desire $\eta$ contains all the predictive information of $x$ with respect to $y$. Therefore we assume that $y$ is independent of $x$ given $\eta$, i.e.,
	\begin{equation}
	p(y,\eta|x)=\Tilde{p}(y|\eta)p(\eta|x)
	\end{equation}
	where the density of $y$ conditioned on $\eta$ is exactly the conditional density the target classifier $f(x)$ built upon. Also, we assume the input lies on a low dimensional manifold, and thus there exists a low dimensional latent variable $z\in\Rm$ conditioned on which $x$ and $\eta$ are independent, i.e.,
	\begin{equation}
	p(\eta, z|x)=p(\eta|z)p(z|x),
	\end{equation}
	where $p(\eta|z)=p(\eta|\mathrm{NN}(z,y))$ and $p(z|x)=p(z|\mathrm{NN}(x,y))$ are Gaussian density parameterized by neural networks. In order to extract class dependent patterns, we make the neural networks also take $y$ as input. Then the conditional density of $y$ given $x$ is
	\begin{equation}
	p(y|x)=\int{p(y|\eta)p(\eta|z)p(z|x)d\eta dz}.
	\end{equation}
	
	\vspace{-3mm}
	\paragraph{Learning} To learn the parameters, We minimize the following objective function,
	\begin{equation}
	\begin{aligned}
	L = \sum_{n=1}^{N}& \mathbb{E}_{p(z_n|x_n,y_n)}\left\{\mathbb{E}_{p(\eta_n|z_n)}\left[\log{\tilde{p}(y_n|\eta_n)}\right]\right\}\\
	&+ \mathrm{KL}\left[p(z_n|x_n,y_n)\Vert N(z_n\vert \bm{0},\bm{I}) \right]\\
	&+ \mathbb{E}_{p(\eta_n|z_n)}[r(\Vert\eta_n\Vert,\epsilon)].
	\end{aligned}
	\label{obj}
	\end{equation}

	All the expectations are taken with respect to reparameterizable densities and are estimated using Monte Carlo integration.
	Note that we do not modify any parameters of the target model $\tilde{p}(y|\eta)$.
	The last term is an additional regularization term, which enforces the norm of $\eta$ does not exceed certain threshold $\epsilon$ in expectation, where $r(\Vert\eta\Vert,\epsilon)=relu(\Vert\eta\Vert-\epsilon)$. We use the $L_\infty$ norm throughout the paper.
	
	The objective (\ref{obj}) consists of three terms. The first term is a lower bound of the log-marginal likelihood and encourages $\eta$ to contain all the predictive information. The second KL divergence term regularizes the encoder $p(z|x)$ towards a standard Gaussian density. The last term is also an additional regularization term, which enforces the norm of $\eta$ does not exceed certain threshold $\epsilon$ in expectation, where $r(\Vert\eta\Vert,\epsilon)=relu(\Vert\eta\Vert-\epsilon)$. We use the $L_\infty$ norm throughout the paper.
	
	\begin{algorithm}[t]
		\label{attack}
		\renewcommand{\algorithmicrequire}{\textbf{Input:}}
		\caption{Adversarial Attacks}
		\begin{algorithmic}
			\REQUIRE a test example $(x^{*},y^{*})$, a target class $y^t$, the target classifier $f$, the encoder $p(z|\mathrm{NN}(x,y))$, the decoder $p(\eta|\mathrm{NN}(z,y))$, the norm bound $\epsilon$, and the maximum number of iterations $K$
			\ENSURE an adversarial example $x^{adv}$
			\STATE $x_{0}^{adv}:=x^{*}$, $n:=0$
			\REPEAT
			\STATE $z_n\sim p(z|\mathrm{NN}(x_{n-1}^{adv},y^*))$; $\eta_n\sim p(\eta|\mathrm{NN}(z_n,y^*))$
			\STATE $x_{n}^{sub}:=\mathrm{Clip}_{x^{*},\epsilon}(x_{n-1}^{adv}-\eta_{n})$
			\STATE Sampled $x^{t}$ from the target class $y^t$
			\STATE $z_{n}^{add}\sim p(z|\mathrm{NN}(x^t,y^t))$; $\eta_{n}^{add}\sim p(\eta|\mathrm{NN}(z_{n}^{add},y^t))$
			\STATE $x_{n}^{adv}:=\mathrm{Clip}_{x^{*},\epsilon}(x_{n}^{sub}+\eta_{n}^{add})$
			\STATE $n:=n+1$
			
			\UNTIL {$f$ misclassifies $x^{adv}$ or $n=K$}
		\end{algorithmic}
	\end{algorithm}
	
	\vspace{-3mm}
	\paragraph{Attack} To perform attack, we replace the predictive patterns $\eta^*$ of a given test example $(x^*,y^*)$ with the patterns $\eta^t$ of some example $(x^t,y^t)$ from a target class $y^t$. In the case of non-targeted attack, we choose to use the class as which the target classifier most probably misclassifies $x^*$, i.e.,
	\begin{equation}
	y^t = \argmax \tilde{p}(y|x^*), \mathrm{s.t.}\, y^t\neq y^*.
	\end{equation}
	To do this, we pass the test example $(x^*,y^*)$ and the target example $(x^t,^t)$ to our model to obtain the patterns $\eta^*$ and $\eta^t$, respectively. Then we replace the patterns in the test input $x^*$ by setting
	\begin{eqnarray}
	\hat{x}^*=x^*-\eta^*+\eta^t.
	\end{eqnarray}
	After every modification, we clip the modified input to enforce the $L_\infty$ norm of its difference with the raw input $x^*$ is below the threshold $\epsilon$. The whole attack procedure is repeated until the target classifier make a mistake or a given maximum number of iterations is reached. We give a summarize of the attack in Algorithm 1.
	
	\begin{figure}
		\begin{subfigure}[b]{0.95\linewidth}
			\centering
			\begin{subfigure}[b]{\linewidth}
    			\begin{subfigure}[b]{0.05\linewidth}
    				\caption*{True}
    				\vspace*{-0.4mm}
    				\rotatebox[origin=t]{90}{\scriptsize Raw}\vspace{0.5\linewidth}
    			\end{subfigure}%
    			\begin{subfigure}[b]{0.1\linewidth}
    				\caption*{0}
    				\vspace*{-2.5mm}
    				\includegraphics[trim={0 0 0 0}, width=\linewidth]{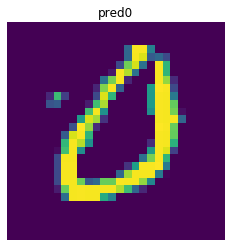}
    			\end{subfigure}%
    			\begin{subfigure}[b]{0.1\linewidth}
    				\caption*{1}
    				\vspace*{-2.5mm}
    				\includegraphics[width=\linewidth]{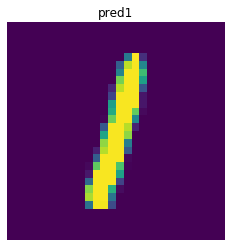}
    			\end{subfigure}%
    			\begin{subfigure}[b]{0.1\linewidth}
    				\caption*{2}
    				\vspace*{-2.5mm}
    				\includegraphics[width=\linewidth]{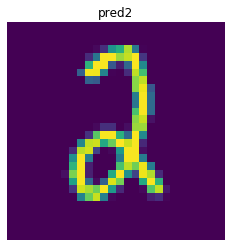}
    			\end{subfigure}%
    			\begin{subfigure}[b]{0.1\linewidth}
    				\caption*{3}
    				\vspace*{-2.5mm}
    				\includegraphics[width=\linewidth]{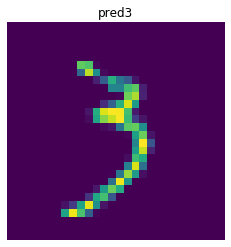}
    			\end{subfigure}%
    			\begin{subfigure}[b]{0.1\linewidth}
    				\caption*{4}
    				\vspace*{-2.5mm}
    				\includegraphics[width=\linewidth]{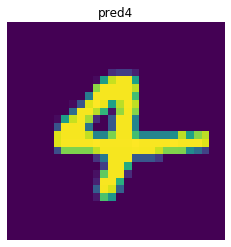}
    			\end{subfigure}%
    			\begin{subfigure}[b]{0.1\linewidth}
    				\caption*{5}
    				\vspace*{-2.5mm}
    				\includegraphics[width=\linewidth]{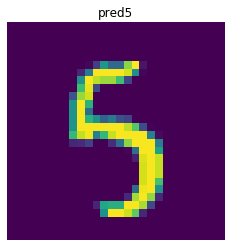}
    			\end{subfigure}%
    			\begin{subfigure}[b]{0.1\linewidth}
    				\caption*{6}
    				\vspace*{-2.5mm}
    				\includegraphics[width=\linewidth]{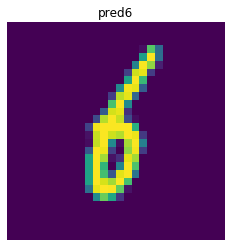}
    			\end{subfigure}%
    			\begin{subfigure}[b]{0.1\linewidth}
    				\caption*{7}
    				\vspace*{-2.5mm}
    				\includegraphics[width=\linewidth]{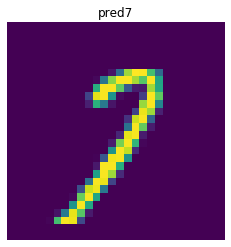}
    			\end{subfigure}%
    			\begin{subfigure}[b]{0.1\linewidth}
    				\caption*{8}
    				\vspace*{-2.5mm}
    				\includegraphics[width=\linewidth]{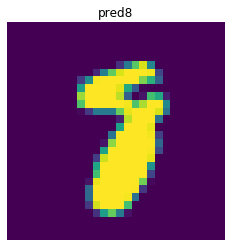}
    			\end{subfigure}%
    			\begin{subfigure}[b]{0.1\linewidth}
    				\caption*{9}
    				\vspace*{-2.5mm}
    				\includegraphics[width=\linewidth]{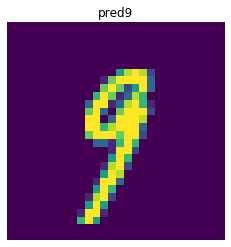}
    			\end{subfigure}\\
    			\begin{subfigure}[b]{0.05\linewidth}
    				\rotatebox[origin=t]{90}{\scriptsize Sub.}\vspace{0.5\linewidth}
    			\end{subfigure}%
    			\begin{subfigure}[b]{0.1\linewidth}
    				\includegraphics[width=\linewidth]{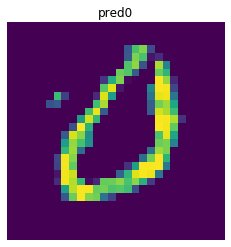}
    			\end{subfigure}%
    			\begin{subfigure}[b]{0.1\linewidth}
    				\includegraphics[width=\linewidth]{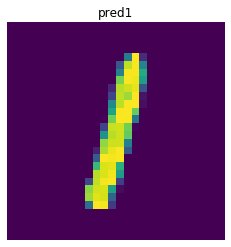}
    			\end{subfigure}%
    			\begin{subfigure}[b]{0.1\linewidth}
    				\includegraphics[width=\linewidth]{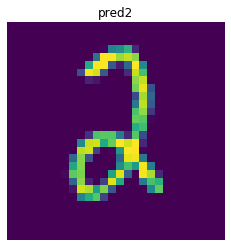}
    			\end{subfigure}%
    			\begin{subfigure}[b]{0.1\linewidth}
    				\includegraphics[width=\linewidth]{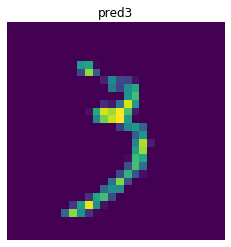}
    			\end{subfigure}%
    			\begin{subfigure}[b]{0.1\linewidth}
    				\includegraphics[width=\linewidth]{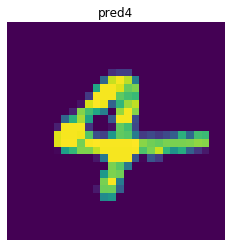}
    			\end{subfigure}%
    			\begin{subfigure}[b]{0.1\linewidth}
    				\includegraphics[width=\linewidth]{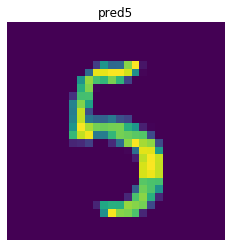}
    			\end{subfigure}%
    			\begin{subfigure}[b]{0.1\linewidth}
    				\includegraphics[width=\linewidth]{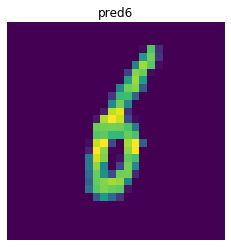}
    			\end{subfigure}%
    			\begin{subfigure}[b]{0.1\linewidth}
    				\includegraphics[width=\linewidth]{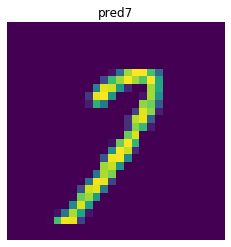}
    			\end{subfigure}%
    			\begin{subfigure}[b]{0.1\linewidth}
    				\includegraphics[width=\linewidth]{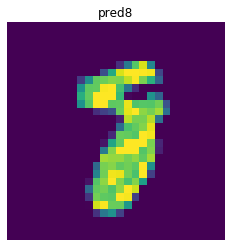}
    			\end{subfigure}%
    			\begin{subfigure}[b]{0.1\linewidth}
    				\includegraphics[width=\linewidth]{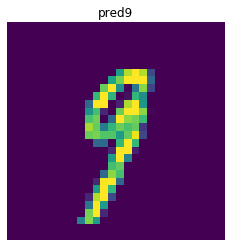}
    			\end{subfigure}\\
    			\begin{subfigure}[b]{0.05\linewidth}
    				\rotatebox[origin=t]{90}{\scriptsize Adv.}\vspace{0.5\linewidth}
    				\caption*{Pred.}
    			\end{subfigure}%
    			\begin{subfigure}[b]{0.1\linewidth}
    				\includegraphics[width=\linewidth]{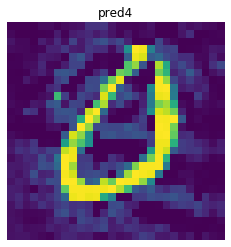}
    				\caption*{4}
    			\end{subfigure}%
    			\begin{subfigure}[b]{0.1\linewidth}
    				\includegraphics[width=\linewidth]{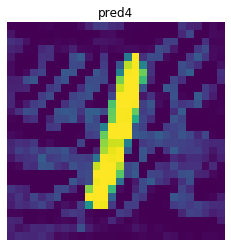}
    				\caption*{4}
    			\end{subfigure}%
    			\begin{subfigure}[b]{0.1\linewidth}
    				\includegraphics[width=\linewidth]{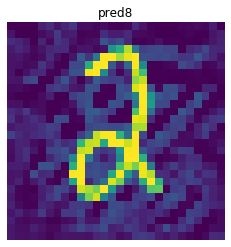}
    				\caption*{8}
    			\end{subfigure}%
    			\begin{subfigure}[b]{0.1\linewidth}
    				\includegraphics[width=\linewidth]{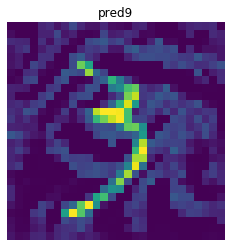}
    				\caption*{9}
    			\end{subfigure}%
    			\begin{subfigure}[b]{0.1\linewidth}
    				\includegraphics[width=\linewidth]{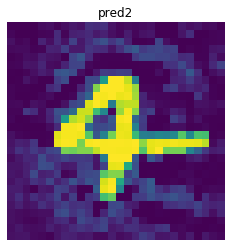}
    				\caption*{2}
    			\end{subfigure}%
    			\begin{subfigure}[b]{0.1\linewidth}
    				\includegraphics[width=\linewidth]{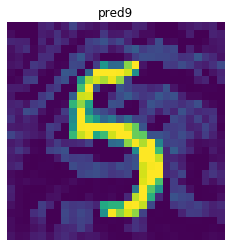}
    				\caption*{9}
    			\end{subfigure}%
    			\begin{subfigure}[b]{0.1\linewidth}
    				\includegraphics[width=\linewidth]{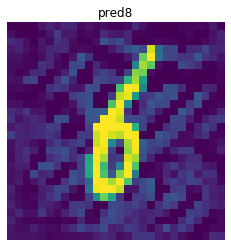}
    				\caption*{8}
    			\end{subfigure}%
    			\begin{subfigure}[b]{0.1\linewidth}
    				\includegraphics[width=\linewidth]{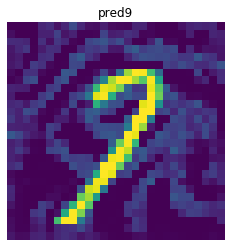}
    				\caption*{9}
    			\end{subfigure}%
    			\begin{subfigure}[b]{0.1\linewidth}
    				\includegraphics[width=\linewidth]{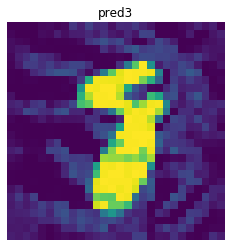}
    				\caption*{3}
    			\end{subfigure}%
    			\begin{subfigure}[b]{0.1\linewidth}
    				\includegraphics[width=\linewidth]{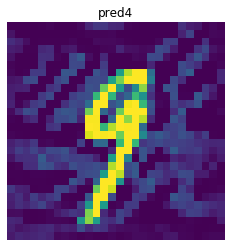}
    				\caption*{4}
    			\end{subfigure}
    		\end{subfigure}\\
    					\begin{subfigure}[b]{\linewidth}
    			\begin{subfigure}[b]{0.05\linewidth}
    				\caption*{True}
    				\vspace*{-0.4mm}
    				\rotatebox[origin=t]{90}{\scriptsize Raw}\vspace{0.5\linewidth}
    			\end{subfigure}%
    			\begin{subfigure}[b]{0.1\linewidth}
    				\caption*{0}
    				\vspace*{-2.5mm}
    				\includegraphics[trim={0 0 0 0}, width=\linewidth]{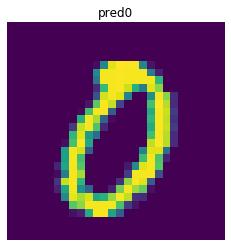}
    			\end{subfigure}%
    			\begin{subfigure}[b]{0.1\linewidth}
    				\caption*{1}
    				\vspace*{-2.5mm}
    				\includegraphics[width=\linewidth]{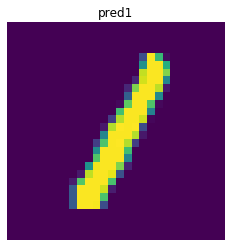}
    			\end{subfigure}%
    			\begin{subfigure}[b]{0.1\linewidth}
    				\caption*{2}
    				\vspace*{-2.5mm}
    				\includegraphics[width=\linewidth]{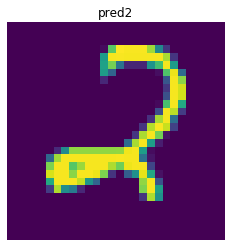}
    			\end{subfigure}%
    			\begin{subfigure}[b]{0.1\linewidth}
    				\caption*{3}
    				\vspace*{-2.5mm}
    				\includegraphics[width=\linewidth]{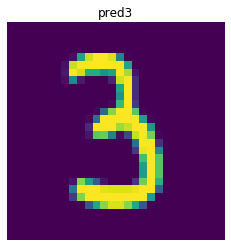}
    			\end{subfigure}%
    			\begin{subfigure}[b]{0.1\linewidth}
    				\caption*{4}
    				\vspace*{-2.5mm}
    				\includegraphics[width=\linewidth]{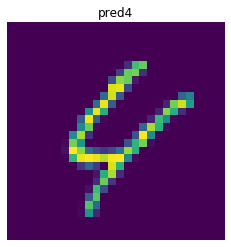}
    			\end{subfigure}%
    			\begin{subfigure}[b]{0.1\linewidth}
    				\caption*{5}
    				\vspace*{-2.5mm}
    				\includegraphics[width=\linewidth]{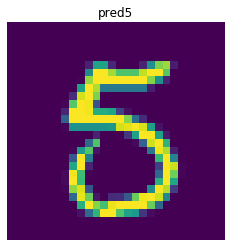}
    			\end{subfigure}%
    			\begin{subfigure}[b]{0.1\linewidth}
    				\caption*{6}
    				\vspace*{-2.5mm}
    				\includegraphics[width=\linewidth]{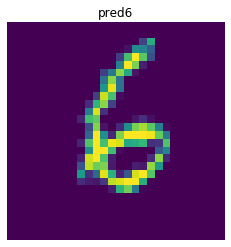}
    			\end{subfigure}%
    			\begin{subfigure}[b]{0.1\linewidth}
    				\caption*{7}
    				\vspace*{-2.5mm}
    				\includegraphics[width=\linewidth]{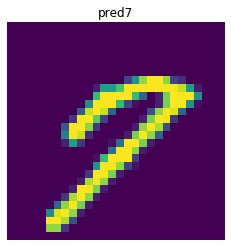}
    			\end{subfigure}%
    			\begin{subfigure}[b]{0.1\linewidth}
    				\caption*{8}
    				\vspace*{-2.5mm}
    				\includegraphics[width=\linewidth]{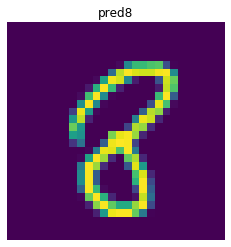}
    			\end{subfigure}%
    			\begin{subfigure}[b]{0.1\linewidth}
    				\caption*{9}
    				\vspace*{-2.5mm}
    				\includegraphics[width=\linewidth]{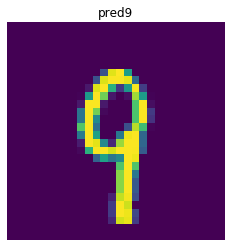}
    			\end{subfigure}\\
    			\begin{subfigure}[b]{0.05\linewidth}
    				\rotatebox[origin=t]{90}{\scriptsize Sub.}\vspace{0.5\linewidth}
    			\end{subfigure}%
    			\begin{subfigure}[b]{0.1\linewidth}
    				\includegraphics[width=\linewidth]{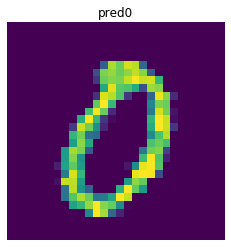}
    			\end{subfigure}%
    			\begin{subfigure}[b]{0.1\linewidth}
    				\includegraphics[width=\linewidth]{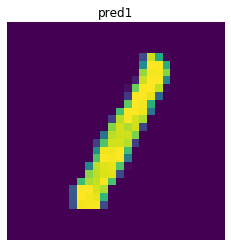}
    			\end{subfigure}%
    			\begin{subfigure}[b]{0.1\linewidth}
    				\includegraphics[width=\linewidth]{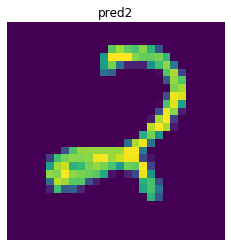}
    			\end{subfigure}%
    			\begin{subfigure}[b]{0.1\linewidth}
    				\includegraphics[width=\linewidth]{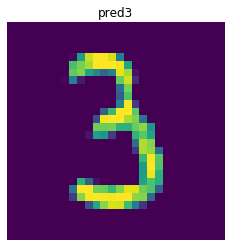}
    			\end{subfigure}%
    			\begin{subfigure}[b]{0.1\linewidth}
    				\includegraphics[width=\linewidth]{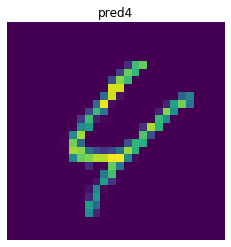}
    			\end{subfigure}%
    			\begin{subfigure}[b]{0.1\linewidth}
    				\includegraphics[width=\linewidth]{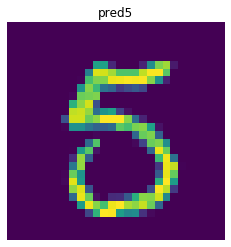}
    			\end{subfigure}%
    			\begin{subfigure}[b]{0.1\linewidth}
    				\includegraphics[width=\linewidth]{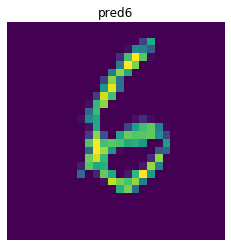}
    			\end{subfigure}%
    			\begin{subfigure}[b]{0.1\linewidth}
    				\includegraphics[width=\linewidth]{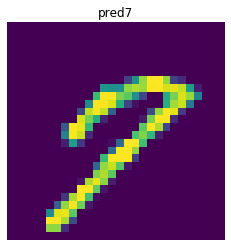}
    			\end{subfigure}%
    			\begin{subfigure}[b]{0.1\linewidth}
    				\includegraphics[width=\linewidth]{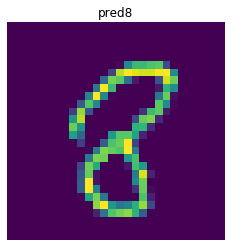}
    			\end{subfigure}%
    			\begin{subfigure}[b]{0.1\linewidth}
    				\includegraphics[width=\linewidth]{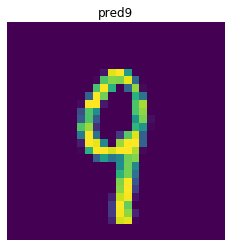}
    			\end{subfigure}\\
    			\begin{subfigure}[b]{0.05\linewidth}
    				\rotatebox[origin=t]{90}{\scriptsize Adv.}\vspace{0.5\linewidth}
    				\caption*{Pred.}
    			\end{subfigure}%
    			\begin{subfigure}[b]{0.1\linewidth}
    				\includegraphics[width=\linewidth]{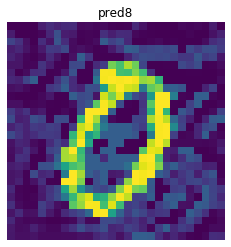}
    				\caption*{4}
    			\end{subfigure}%
    			\begin{subfigure}[b]{0.1\linewidth}
    				\includegraphics[width=\linewidth]{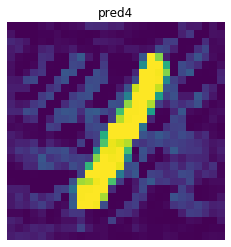}
    				\caption*{4}
    			\end{subfigure}%
    			\begin{subfigure}[b]{0.1\linewidth}
    				\includegraphics[width=\linewidth]{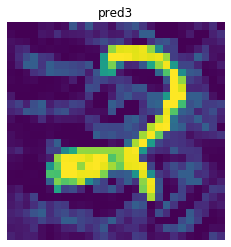}
    				\caption*{8}
    			\end{subfigure}%
    			\begin{subfigure}[b]{0.1\linewidth}
    				\includegraphics[width=\linewidth]{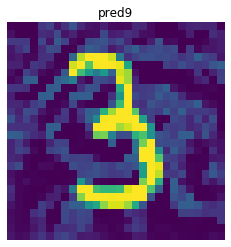}
    				\caption*{9}
    			\end{subfigure}%
    			\begin{subfigure}[b]{0.1\linewidth}
    				\includegraphics[width=\linewidth]{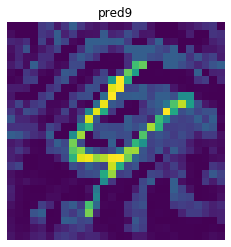}
    				\caption*{2}
    			\end{subfigure}%
    			\begin{subfigure}[b]{0.1\linewidth}
    				\includegraphics[width=\linewidth]{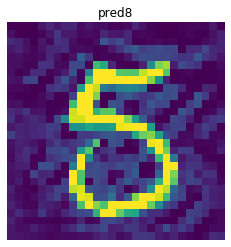}
    				\caption*{9}
    			\end{subfigure}%
    			\begin{subfigure}[b]{0.1\linewidth}
    				\includegraphics[width=\linewidth]{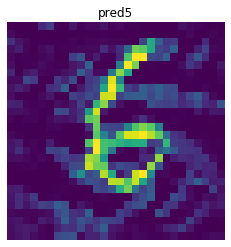}
    				\caption*{8}
    			\end{subfigure}%
    			\begin{subfigure}[b]{0.1\linewidth}
    				\includegraphics[width=\linewidth]{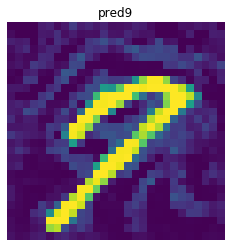}
    				\caption*{9}
    			\end{subfigure}%
    			\begin{subfigure}[b]{0.1\linewidth}
    				\includegraphics[width=\linewidth]{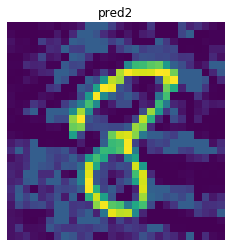}
    				\caption*{3}
    			\end{subfigure}%
    			\begin{subfigure}[b]{0.1\linewidth}
    				\includegraphics[width=\linewidth]{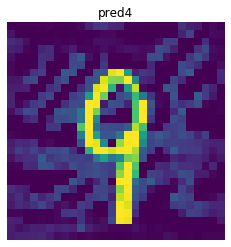}
    				\caption*{4}
    			\end{subfigure}
    		\end{subfigure}\\
		\end{subfigure}\hfill\\
		\vspace{-6mm}
		\caption{A visualization of the attack procedure. The three rows show the raw image, the image with generated patterns subtracted, and the adversarial image (the image with patterns replaced). The true labels and the predictions given by the target classifier is shown at the top and the bottom, respectively.}
		\label{visual}
	\end{figure}
	
	\section{Experiments}
	\subsection{Configurations}
	\paragraph{Hyperparameters and implementation details} The architecture of the proposed model: The encoder has 2 layers with kernel sizes= [5, 5], strides= [1, 2, 2, 1] and feature map sizes= [784,256]. The Decoder architecture has also 2 layers with kernel sizes= [5, 5], strides= [1, 2, 2, 1] and feature map sizes= [256,784]. The first layers have RELU activation functions, the last layer is a full connection layer, and all layers except the last one use Batch Normalization. We trained the proposed model with the Adam optimizer. We tuned the dimension $L$ of the latent space (ending up with L = 30); started with a high weight for the KL-divergence term at the beginning of training (which was gradually decreased from a factor of 10 to 1 over 50 epochs); 

    \paragraph{Hyperparameters for the attack model} We set the adversarial perturbation size $\epsilon_{L_\infty} = 0.3$ as a threshold. We set the number of the iteration $t=30$ as a threshold. 

    \paragraph{Hyperparameters for the target model} The architecture of the CNNs has kernel sizes = [5, 4, 3, 5], strides = [1, 2, 2, 1], and feature map sizes = [20, 70, 256, 10]. All layers use ELU activation functions and all layers except the last one apply Batch Normalization. The CNNs are both trained on the cross entropy loss with the Adam optimizer. The parameters maximizing the test cross entropy.

	\subsection{Evaluation}
	We evaluate our model on MNIST. We apply our model to generate adversarial examples to a convolutional neural network (CNN). We generate adversarial examples under an $L_\infty$ bound of 0.3. Table \ref{cnn_sar} shows the successful attack rate and the original accuracy of CNN.
	\begin{table}
		\centering
		\begin{tabular}{l|l}
			\hline
			Model                    & CNN      \\ 
			\hline
			Accuracy             & 99.35\%  \\ 
			\hline
			Attack Success Rate & 94.14\%\\
			\hline
		\end{tabular}
		\vspace{-3mm}
		\caption{Model Accuracy and Attack Success Rate}
		\label{cnn_sar}
	\end{table}
	
	To demonstrate that our model is capable of learning predictive patterns, we let the target model make prediction basing on the generated patterns solely, and it achieve $100\%$ accuracy on the test set. We report the average $L_\infty$ norm of the generated patterns from each class in Table $\ref{norm}$. It can be seen that the norm of generated patterns are below the threshold on average.
	\begin{table}
	\centering
	\begin{tabular}{llllll} 
	\hline
	Class     & 0                         & 1                         & 2                         & 3                         & 4                          \\ 
	\hline
	Avg. Norm & \multicolumn{1}{r}{0.291} & \multicolumn{1}{r}{0.178} & \multicolumn{1}{r}{0.284} & \multicolumn{1}{r}{0.289} & \multicolumn{1}{r}{0.294}  \\ 
	\hline
	\hline
	Class     & 5                         & 6                         & 7                         & 8                         & 9                          \\ 
	\hline
	Avg. Norm & 0.275                     & 0.289                     & 0.268                     & 0.293                     & 0.289                      \\
	\hline
	\end{tabular}
	
	\caption{Average Norm of Generated Patterns}
	\vspace{-5mm}
	\label{norm}
	\end{table}
	
\begin{figure}
		\centering
		    \begin{subfigure}[b]{0.9\linewidth}
		        \includegraphics[width=\linewidth]{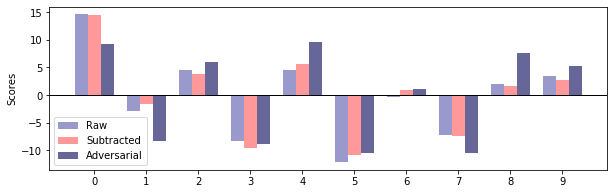}
		    \end{subfigure}\\
		    \begin{subfigure}[b]{0.9\linewidth}
		        \includegraphics[width=\linewidth]{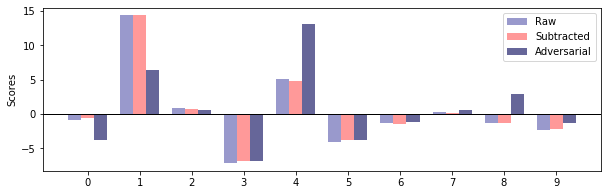}
		    \end{subfigure}\\
		    \begin{subfigure}[b]{0.9\linewidth}
		        \includegraphics[width=\linewidth]{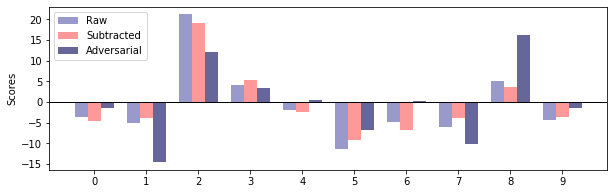}
		    \end{subfigure}\\
		\begin{subfigure}[b]{0.9\linewidth}
			\includegraphics[width=\linewidth]{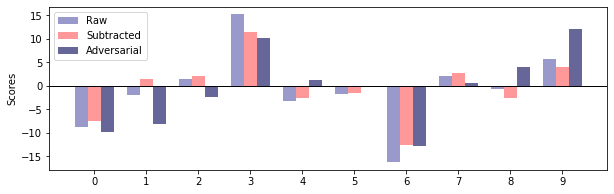}
		\end{subfigure}\\
		    \begin{subfigure}[b]{0.9\linewidth}
		        \includegraphics[width=\linewidth]{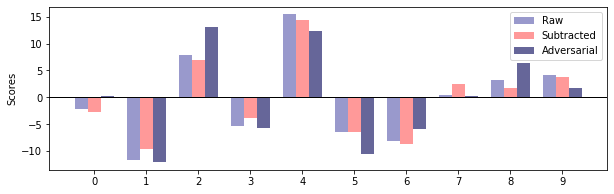}
		    \end{subfigure}\\
		\begin{subfigure}[b]{0.9\linewidth}
			\includegraphics[width=\linewidth]{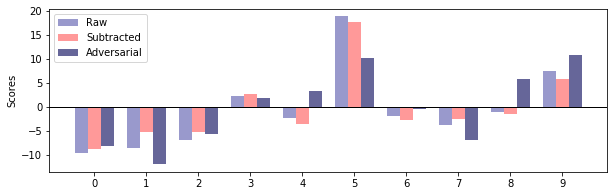}
		\end{subfigure}\\
		\begin{subfigure}[b]{0.9\linewidth}
			\includegraphics[width=\linewidth]{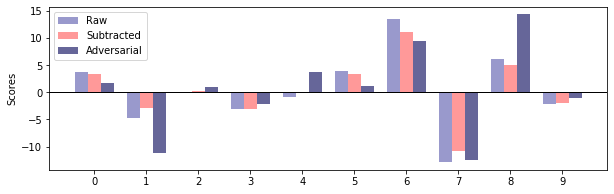}
		\end{subfigure}\\
		    \begin{subfigure}[b]{0.9\linewidth}
		        \includegraphics[width=\linewidth]{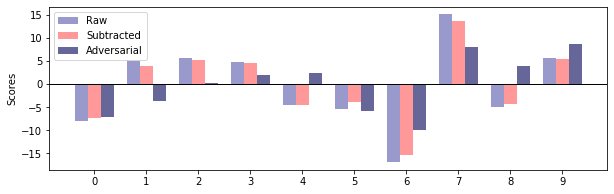}
		    \end{subfigure}\\
		    \begin{subfigure}[b]{0.9\linewidth}
		        \includegraphics[width=\linewidth]{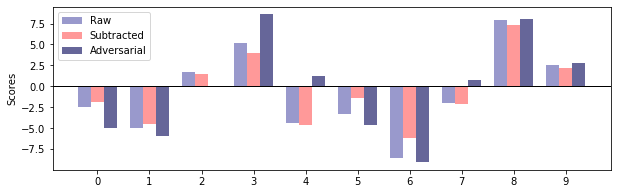}
		    \end{subfigure}\\
		    \begin{subfigure}[b]{0.9\linewidth}
		        \includegraphics[width=\linewidth]{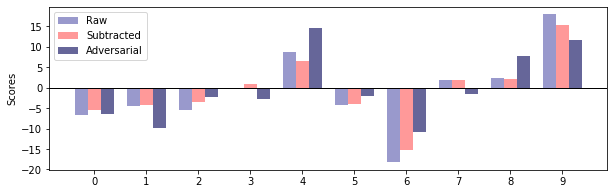}
		    \end{subfigure}\\
		\caption{Changes in scores the target classifier assigning to different class during attack.}
		\label{fig:vary_scores}
	\end{figure}

	In Figure \ref{visual}, we give a visualization of the images under the attack procedure. We also plot changes in scores the target classifier assigning to different classes during attack in Figure \ref{fig:vary_scores}. We can see that the score of the true label given by the target classifier decreases after the generated patterns are subtracted from the raw input, and the score of the target class increases significantly after the patterns of the input are replaced.

	\bibliographystyle{aaai}
	\bibliography{ref}
\end{document}